\PassOptionsToPackage{unicode}{hyperref}
\PassOptionsToPackage{hyphens}{url}
\documentclass[
]{article}
\usepackage{xcolor}
\usepackage{amsmath,amssymb}
\usepackage[margin=1in]{geometry}
\usepackage{iftex}
\ifPDFTeX
  \usepackage[T1]{fontenc}
  \usepackage[utf8]{inputenc}
  \usepackage{textcomp} 
\else 
  \usepackage{unicode-math} 
  \defaultfontfeatures{Scale=MatchLowercase}
  \defaultfontfeatures[\rmfamily]{Ligatures=TeX,Scale=1}
\fi
\usepackage{lmodern}
\ifPDFTeX\else
\fi
\IfFileExists{upquote.sty}{\usepackage{upquote}}{}
\IfFileExists{microtype.sty}{
  \usepackage[]{microtype}
  \UseMicrotypeSet[protrusion]{basicmath} 
}{}
\makeatletter
\@ifundefined{KOMAClassName}{
  \IfFileExists{parskip.sty}{%
    \usepackage{parskip}
  }{
    \setlength{\parindent}{0pt}
    \setlength{\parskip}{6pt plus 2pt minus 1pt}}
}{
  \KOMAoptions{parskip=half}}
\makeatother
\usepackage{graphicx}
\makeatletter
\newsavebox\pandoc@box
\newcommand*\pandocbounded[1]{
  \sbox\pandoc@box{#1}%
  \Gscale@div\@tempa{\textheight}{\dimexpr\ht\pandoc@box+\dp\pandoc@box\relax}%
  \Gscale@div\@tempb{\linewidth}{\wd\pandoc@box}%
  \ifdim\@tempb\p@<\@tempa\p@\let\@tempa\@tempb\fi
  \ifdim\@tempa\p@<\p@\scalebox{\@tempa}{\usebox\pandoc@box}%
  \else\usebox{\pandoc@box}%
  \fi%
}
\def\fps@figure{htbp}
\makeatother
\providecommand{\tightlist}{%
  \setlength{\itemsep}{0pt}\setlength{\parskip}{0pt}}
\usepackage{bookmark}
\IfFileExists{xurl.sty}{\usepackage{xurl}}{} 
\hypersetup{
  hidelinks,
  pdfcreator={LaTeX via pandoc}}

\title{\textbf{Leverage Laws: A Per-Task Framework for Human-Agent Collaboration}}
\author{Stan Loosmore \\ \texttt{loosmore@usc.edu}}
\date{April 2026}

\begin{document}

\maketitle

\subsection*{Abstract}

We propose a per-task leverage ratio for human-agent collaboration:
human work displaced by an agent, divided by the human time required to
specify the task, resolve mid-run interrupts, and review the result. The
denominator decomposes into three channels through which a conserved
per-task information requirement must flow, each with its own time-cost
scalar. We show that information density itself is directional and
bounded by separate ceilings on human-to-agent and agent-to-human flow,
and that the asymptotic behavior of leverage decomposes into two scaling
axes (capability and memory) with a non-zero floor on the planning term
set by irreducible task novelty bounded by human throughput. We extend
this per-task analysis to a windowed leverage measure that accommodates
recurring tasks, spawned subtasks, and amortized system-design
investment. The per-task ceiling does not bind the windowed measure,
though both remain bounded: \(L_{\text{task}}\) by per-task novelty,
\(L_{\text{window}}\) by the stock of accumulated planning investment
that pays out within the window. The framework operationalizes aspects
of earlier qualitative work on supervisory control [1],
common ground [2], and mixed-initiative interaction
[3] within a single normative ratio, and produces a list of
testable empirical questions including a Popper-grade falsification
protocol for the directional asymmetry claim.

\section{1. Introduction}\label{introduction}

The productivity literature on human-AI collaboration has converged on
three well-explored claims: that AI improves output [4, 5],
that the productivity frontier is jagged across tasks [6],
and that autonomous time horizons are growing [7]. What this literature does not provide
is a unit of analysis. There is no per-task ratio that lets a
practitioner answer the operational question of where to invest the next
hour of work to increase the rate at which agents displace human labor.

We propose such a ratio. The contribution is not the recognition that
human-agent collaboration has a leverage structure; that observation is
implicit across the supervisory control and common ground literatures.
The contribution is a unified formulation that connects three previously
disjoint claims:

\begin{enumerate}
\def\labelenumi{\arabic{enumi}.}
\tightlist
\item
  The cost of supervising autonomous systems can be decomposed into
  measurable channels [1].
\item
  The information required to specify a task decreases with shared
  context between collaborators [2].
\item
  The decision to interrupt a human or autonomous process is a
  cost-benefit calculation [3].
\end{enumerate}

We show these are three views of one ratio, with a tractable asymptotic
analysis and one Popper-grade falsifiable prediction concerning
directional density and phase-specific time savings.

\section{2. The Base Formula}\label{the-base-formula}

Define per-task leverage as:

\[
L_{\text{task}} \;=\; \frac{H_{\text{displaced}}}{t_{\text{planning}} \;+\; \sum_{j} t_{\text{interrupt}_j} \;+\; t_{\text{review}}}
\]

Here \(L_{\text{task}}\) is the dimensionless leverage of a single task
(human-hours displaced per human-hour spent), \(H_{\text{displaced}}\)
is the human work the agent removes from the operator's plate,
\(t_{\text{planning}}\) is the human time spent specifying the task
before dispatch, \(t_{\text{interrupt}_j}\) is the human time spent
resolving the \(j\)-th interrupt mid-run, and \(t_{\text{review}}\) is
the human time spent verifying the agent's output. All time terms are in
hours.

The numerator is governed by agent capability. The denominator is
governed by the structure of the human-agent exchange. The two are
\emph{first-order decoupled}: investments in models and infrastructure
raise the numerator without directly affecting the denominator, and
investments in workflow design lower the denominator without changing
capability. At second order, the two coevolve. A more capable model
reduces interrupt frequency by needing fewer clarifications; richer
tools both increase \(H_{\text{displaced}}\) and shorten review by
producing more verifiable output. The decoupling is an engineering
separability rather than a strict structural independence: the levers
can be funded as distinct programs, but a capability improvement raises
every term in the equation, and the framework identifies which terms
move first under which interventions.

Each denominator term is an \emph{exchange}: a bounded interaction in
which information is transferred between the human and the agent.

\section{3. Information Density}\label{information-density}

\subsection{3.1 Definition}\label{definition}

Define information density:

\[
\rho \;=\; \frac{I_{\text{conveyed}}}{t_{\text{exchange}}}
\]

Here \(\rho\) is the information density of an exchange (in bits per
hour), \(I_{\text{conveyed}}\) is the useful information transferred (in
bits), and \(t_{\text{exchange}}\) is the wall-clock duration of the
exchange (in hours).

We use Shannon bits as the unit; the qualifier ``useful'' is
load-bearing, and a formal treatment of which bits are task-relevant is
deferred to §7.

\subsection{3.2 Directional
Decomposition}\label{directional-decomposition}

Information density is asymmetric: human-to-agent flow and
agent-to-human flow have different bottlenecks. Let \(\rho_{\text{in}}\)
denote the human-to-agent rate, bottlenecked by human output (speech,
with asymptote near 20 bits/sec; typing; intent compression). Let
\(\rho_{\text{out}}\) denote the agent-to-human rate, bottlenecked by
human input (reading, listening, visual parsing); the agent's choice of
modality determines which channel is used.

For an exchange in which fraction \(\alpha \in [0, 1]\) of the
information flows human-to-agent, the time required to move \(I\) total
bits is:

\[
t_{\text{exchange}}(I, \alpha) \;=\; \frac{\alpha \cdot I}{\rho_{\text{in}}} \;+\; \frac{(1 - \alpha) \cdot I}{\rho_{\text{out}}}
\]

The effective density of the exchange is the harmonic mean of
\(\rho_{\text{in}}\) and \(\rho_{\text{out}}\) weighted by the
information shares:

\[
\rho_{\text{eff}}(\alpha) \;=\; \frac{1}{\alpha / \rho_{\text{in}} \;+\; (1 - \alpha) / \rho_{\text{out}}}
\]

This form is correct because rates combine in inverse when both
bottlenecks gate the same exchange.

\begin{figure}
\centering
\pandocbounded{\includegraphics[keepaspectratio,alt={Effective density as the harmonic mean of two directional ceilings, with the three channels marked along the curve}]{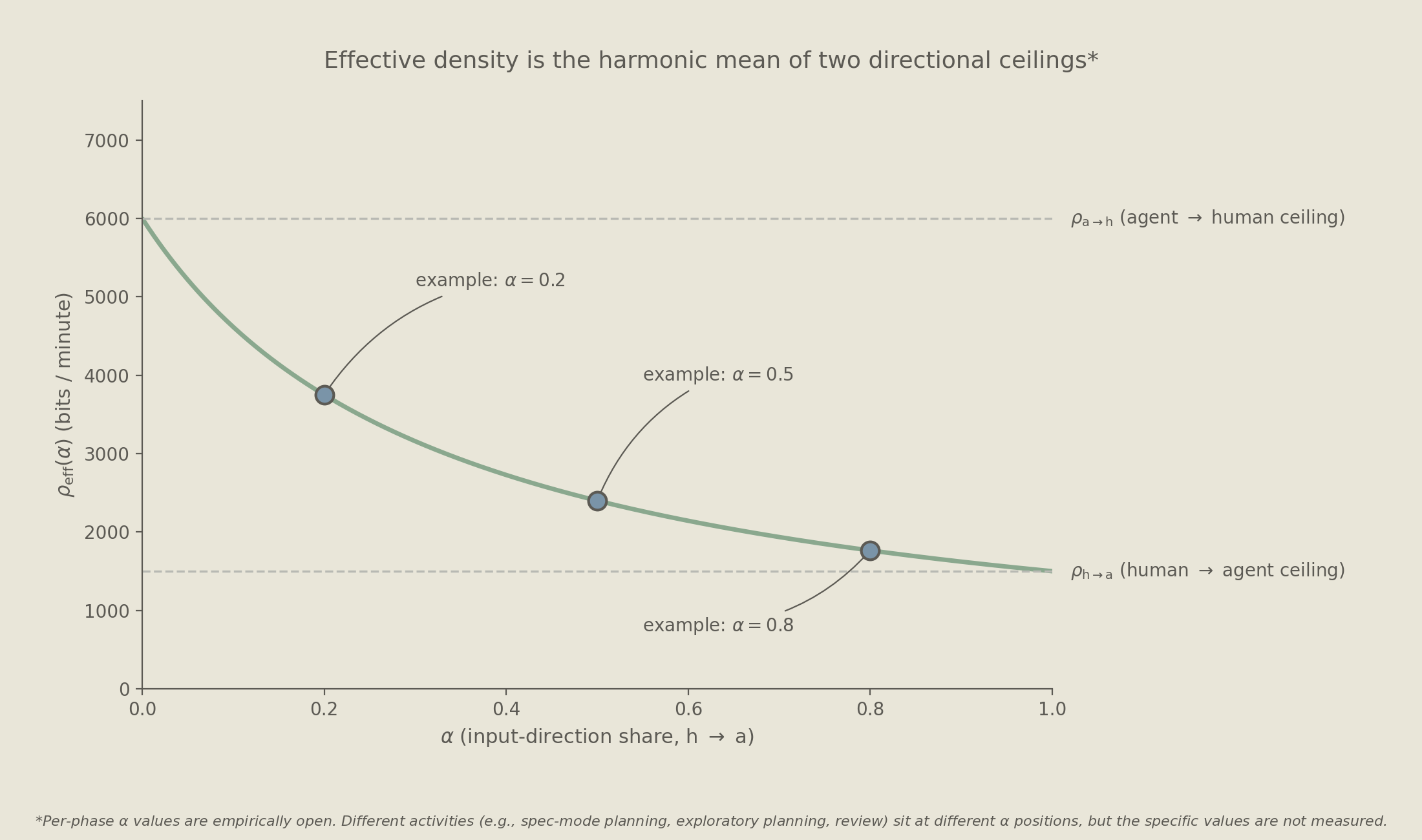}}
\caption{Effective density as the harmonic mean of two directional
ceilings, with the three channels marked along the curve}
\end{figure}

\subsection{3.3 Channel-Specific
Mixtures}\label{channel-specific-mixtures}

Each term in the denominator of \(L_{\text{task}}\) has its own
characteristic \(\alpha\), and the value depends on the activity within
the phase, not on the phase label alone. We do not commit to specific
values for any phase; the values below are illustrative orderings, not
measurements.

\begin{itemize}
\tightlist
\item
  \textbf{Planning} depends on planning mode. \emph{Spec-mode planning}
  (the human knows what they want and is conveying intent) tends to be
  input-heavy: \(\alpha_p\) is high. \emph{Exploratory planning} (the
  human is using the agent to brainstorm and figure out what they want)
  is more bidirectional: \(\alpha_p\) is closer to balanced.
\item
  \textbf{Interrupt} is typically bidirectional. The agent communicates
  state (\(\rho_{\text{out}}\)); the human resolves it
  (\(\rho_{\text{in}}\)). \(\alpha_{i,j}\) tends toward the middle of
  the range.
\item
  \textbf{Review} is typically output-heavy: agent-to-human result
  transfer, with some human correction. \(\alpha_r\) tends to be low,
  higher when the human pushes back substantively.
\end{itemize}

The structural claim is that these three channels span a \emph{range} of
\(\alpha\), and that workflows in which all channels collapse to the
same \(\alpha\) are degenerate. The specific values are empirically open
and likely activity-dependent (open problem 9). The operational
consequence holds for any \(\alpha\)-ordering: a \(\rho_{\text{in}}\) improvement
(e.g., voice input) primarily helps phases with higher \(\alpha\), while
a \(\rho_{\text{out}}\) improvement (e.g., a structured trace dashboard)
primarily helps phases with lower \(\alpha\).

\subsection{3.4 Asymmetric Ceilings}\label{asymmetric-ceilings}

\(\rho_{\text{in}}\) and \(\rho_{\text{out}}\) have different headroom.
\(\rho_{\text{in}}\) is bounded above by human output rate, with speech
as the natural asymptote near \(\sim 20\) bits/sec.~Tools like
voice-to-text are already approaching this ceiling; further gains
require either higher information density per spoken token (better
intent compression) or new modalities (e.g., direct neural interfaces)
that we treat as out of scope.

\(\rho_{\text{out}}\) has substantially more headroom because the agent
can choose its output modality. A well-designed dashboard or diagram
conveys orders of magnitude more bits per second than an equivalent
prose summary. The ceiling on \(\rho_{\text{out}}\) is set by the
human's parallel visual processing capacity, which is poorly
characterized but clearly higher than read-aloud rates.

\(\rho_{\text{out}}\) has more headroom for improvement than
\(\rho_{\text{in}}\), but improvements to \(\rho_{\text{out}}\) cannot
rescue exchanges bottlenecked on \(\rho_{\text{in}}\), and vice versa.
The two are separate investment levers.

\section{4. Information Conservation}\label{information-conservation}

\subsection{4.1 The Conservation Law}\label{the-conservation-law}

Every task has a total information requirement that must be conveyed for
the agent to succeed. Call it \(I_{\text{task}}\). The operator does not
choose whether to pay it. They only choose where:

\[
I_{\text{task}} \;\leq\; I_{\text{planning}} \;+\; \sum_{j} I_{\text{interrupt}_j} \;+\; I_{\text{review}}
\]

The inequality is non-strict because workflows can convey the same
information twice (redundancy) or convey more than \(I_{\text{task}}\)
requires (over-specification). Equality holds when the workflow conveys
no redundant or supra-task bits; the distribution across channels is not
unique.

The three channels trade off: sparse planning forces information into
interrupts; insufficient interrupt resolution forces it into review.
This is the formal expression of the intuition that under-specifying a
task pushes its cost into later, more expensive phases.

\subsection{4.2 Channel Cost Scalars}\label{channel-cost-scalars}

Each channel has its own time cost per unit of information. Combining
the directional decomposition (§3.2) with the conservation law:

\[
t_{\text{planning}} \;=\; \frac{c_p \cdot I_{\text{planning}}}{\rho_{\text{eff}}(\alpha_p)} \qquad t_{\text{interrupt}_j} \;=\; \frac{c_{i,j} \cdot I_{\text{interrupt}_j}}{\rho_{\text{eff}}(\alpha_{i,j})} \qquad t_{\text{review}} \;=\; \frac{c_r \cdot I_{\text{review}}}{\rho_{\text{eff}}(\alpha_r)}
\]

Here \(c_p\), \(c_{i,j}\), and \(c_r\) are dimensionless cost scalars
for planning, the \(j\)-th interrupt, and review respectively. Each
interrupt carries its own context-switch overhead, hence the
per-interrupt indexing.

The scalars capture overhead beyond the raw information transfer:
context-switch costs, rework induced by late-arriving information, and
re-orientation time when the workflow has stalled.

\subsection{4.3 The Substituted Form}\label{the-substituted-form}

Combining §3 and §4:

\[
L_{\text{task}} \;=\; \frac{H_{\text{displaced}}}{\sum_{\text{term}} c_{\text{term}} \cdot I_{\text{term}} / \rho_{\text{eff}}(\alpha_{\text{term}})}
\]

Every variable on the right-hand side maps to a separate engineering
target:

\begin{itemize}
\tightlist
\item
  \(H_{\text{displaced}}\) --- model capability and access to skills,
  tools, infrastructure.
\item
  \(c_{\text{term}}\) --- workflow design (specification quality,
  interrupt protocol, review automation).
\item
  \(I_{\text{term}}\) distribution --- once \(c_{\text{term}}\) have
  been measured for the workflow, route information into channels where
  they are smallest.
\item
  \(\rho_{\text{in}}\) --- human input modality (voice, intent
  compression).
\item
  \(\rho_{\text{out}}\) --- agent output modality (visualization,
  structured display).
\item
  \(\alpha_{\text{term}}\) --- workflow protocol (how much each phase
  relies on each direction).
\end{itemize}

\section{5. Limits and Growth}\label{limits-and-growth}

\subsection{5.1 Numerator Growth}\label{numerator-growth}

Capability grows without bound:

\[
\lim_{\text{capability} \to \infty} H_{\text{displaced}} \;\to\; \infty
\]

Capability here aggregates model intelligence, skill availability, tool
access, and infrastructure. This is the well-explored axis [7].

\subsection{5.2 Memory and Density
Growth}\label{memory-and-density-growth}

Information density is a function of accumulated shared memory:

\[
\rho_{\text{in}} \;=\; \rho_{\text{in}}(M), \quad \rho_{\text{out}} \;=\; \rho_{\text{out}}(M), \quad \frac{\partial \rho}{\partial M} > 0
\]

where \(M\) is the operator's accumulated shared context with the agent:
skills, prior dispatches, project memory, agent notes, and any
persistent representation that compresses future exchanges. Memory
raises density because each unit of new communication carries more
meaning when both sides already share context. The qualitative claim
that grounding cost decreases with shared context is established in
common ground theory [2]; \(\rho(M)\) is one
quantitative operationalization.

We assume reliable memory in the analysis. When the agent is uncertain
about a piece of prior context, it can request clarification, which adds
a small increment to \(t_{\text{planning}}\) at the standard cost.
Memory quality (staleness, ambiguity, miscalibration) is a separate
engineering problem from memory accumulation, and is decoupled from the
framework presented here.

The two density axes scale at different rates. \(\rho_{\text{in}}(M)\)
is bounded by human output ceilings and saturates relatively quickly.
\(\rho_{\text{out}}(M)\) is bounded only by the agent's modality choices
and human visual parsing, and continues to scale as agents learn better
output formats.

\subsection{5.3 Channel-Wise Limits}\label{channel-wise-limits}

Two of the three denominator terms collapse to zero in the joint limit:

\[
\lim_{M \to \infty} \sum_{j} t_{\text{interrupt}_j} \;\to\; 0
\]

Every interrupt is information the agent needed but did not have at
dispatch. Under perfect specification, there is nothing left to ask.

\[
\lim_{M \to \infty} t_{\text{review}} \;\to\; 0
\]

Review collapses as verification moves into the system. The verifier
need not be the same agent that produced the output: multi-agent
orchestration, automated checks against ground truth, and ensemble
cross-review absorb what a human used to perform manually.

The planning term approaches a non-zero floor:

\[
\lim_{M \to \infty} t_{\text{planning}} \;=\; \frac{I_{\text{novel}}}{\rho_{\text{eff}}(\alpha_p, M)} \;>\; 0
\]

where \(I_{\text{novel}}\) is the irreducible new information specific
to the task that no memory could anticipate. This floor is the
irreducibility claim: planning cannot disappear, only compress.

\begin{figure}
\centering
\pandocbounded{\includegraphics[keepaspectratio,alt={Each time term as the system matures: planning floors at a non-zero value, interrupts and review go to zero}]{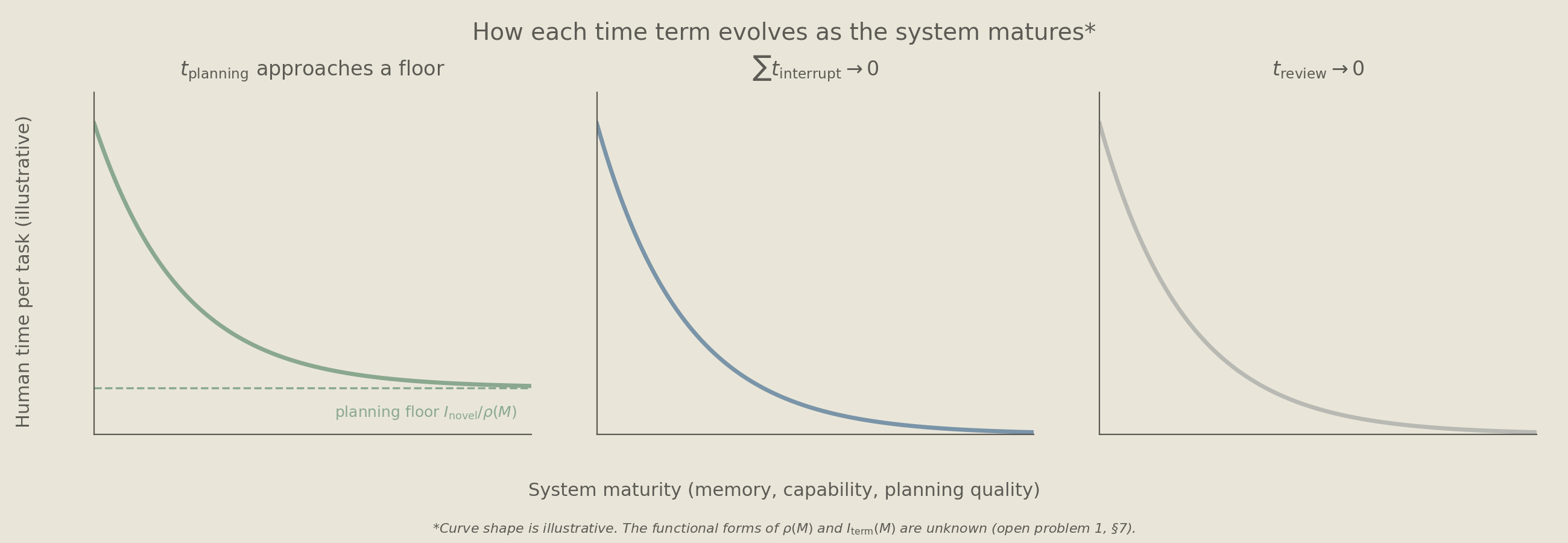}}
\caption{Each time term as the system matures: planning floors at
\(I_{\text{novel}} / \rho(M)\), interrupts and review go to zero}
\end{figure}

\subsection{5.4 Joint Limit}\label{joint-limit}

Combining:

\[
\lim_{M, \text{capability} \to \infty} L_{\text{task}} \;=\; \frac{H_{\text{displaced}}}{I_{\text{novel}} / \rho_{\text{eff}}(\alpha_p, M)}
\]

For a single bounded task, this limit is finite. Both
\(H_{\text{displaced}}\) and \(I_{\text{novel}}\) are properties of the
task itself: the human work the task would have demanded, and the
irreducible novel information the task introduces relative to memory.
Both are task-fixed, independent of session length. The bound on
per-task leverage in the limit is therefore:

\[
L_{\text{task}}^{\max} \;=\; \frac{H_{\text{displaced}} \cdot \rho_{\text{in}}^{\max}}{I_{\text{novel}}}
\]

Per-task leverage approaches a ceiling set by the ratio of the task's
displaced work to its novelty, scaled by human input bandwidth. The
ceiling is finite because the task is finite, not because the planning
session is.

The binding constraint on this ceiling is \(\rho_{\text{in}}\), since
\(\rho_{\text{eff}}\) at planning depends primarily on the input
direction (\(\alpha_p \approx 0.8\)). Investments that raise
\(\rho_{\text{in}}\) (better input modalities, intent compression) are
the only investments that raise the per-task ceiling.

This ceiling is \emph{per-task}. When tasks recur, spawn other tasks, or
share planning investment through system-design upstream, the ceiling
amortizes and the unit of analysis shifts from task to time. §6 develops
the amortization argument.

\section{6. Beyond the Single Task}\label{beyond-the-single-task}

The per-task ceiling in §5.4 binds only for one-off tasks. Real
workflows aggregate many tasks across time, some recurring, some
spawning others, and the right unit of analysis at this scale is
leverage per time window rather than leverage per task.

Define leverage over a window of operator time:

\[
L_{\text{window}}(T) \;=\; \frac{\sum_k H_{\text{displaced},k}}{T_{\text{operator}}}
\]

where the sum is over all tasks \(k\) completed in the window and
\(T_{\text{operator}}\) is the operator's hours actually spent within
the window. This unit handles three regimes that \(L_{\text{task}}\)
does not:

\begin{enumerate}
\def\labelenumi{\arabic{enumi}.}
\item
  \textbf{Recurring tasks}. A task scheduled to run \(N\) times pays its
  planning cost once. Per-run effective planning time is
  \(t_{\text{planning}} / N\) and approaches zero as \(N\) grows.
  Combined with §5.3 (interrupts and review collapse in the joint
  limit), the per-run leverage of a stationary recurring task grows
  without per-task ceiling. In non-stationary environments, a
  maintenance term \(t_{\text{maint}}(\tau)\) enters the denominator and
  bounds per-run leverage at the rate of environmental drift (schema
  changes, dependency upgrades, world drift).
\item
  \textbf{Spawned tasks}. Children of a parent task inherit the parent's
  accumulated \(M\), which raises \(\rho_{\text{eff}}\) on every child
  dispatch and reduces, but does not eliminate, per-child
  \(t_{\text{planning}}\). The parent absorbs the \(I_{\text{novel}}\)
  of the decomposition itself; children pay only the residual novelty
  their subtask introduces. The numerator of \(L_{\text{window}}\)
  accumulates over the tree while the operator's per-child planning cost
  is reduced by inherited memory rather than removed by it.
\item
  \textbf{System design upstream}. Skills, templates, agent memory, and
  other persistent representations move what would have been per-task
  novelty into one-time investments that amortize across all future
  tasks. From the perspective of \(L_{\text{window}}\), system-design
  hours are paid in the window they happen and pay back in every
  subsequent window.
\end{enumerate}

The per-task floor in §5.3 is not contradicted by these amortization
regimes. The floor is paid in human-hours per unit of novel intent;
amortization redistributes that cost across the displaced work it
enables. \(L_{\text{task}}\) is bounded; \(L_{\text{window}}\) has no
per-task ceiling but is itself bounded by the stock of accumulated
planning investment that pays out in the window, an investment paid in
operator-hours during earlier windows. The two measures answer different
questions and hold simultaneously.

A second ratio supports the windowed framing:

\[
E_{\text{task}} \;=\; \frac{H_{\text{displaced}}}{t_{\text{agent}}}
\]

where \(t_{\text{agent}}\) is the wall-clock duration of the agent's
run. \(E_{\text{task}}\) does not enter \(L_{\text{task}}\), but it
governs how many tasks the operator can run in parallel within a window:
low \(t_{\text{agent}}\) allows pipelining, high \(t_{\text{agent}}\)
forces serial scheduling.

When tasks have dependencies, \(L_{\text{window}}\) is no longer a sum
of independent contributions. It becomes a scheduling problem: maximize
\(L_{\text{window}}\) subject to the constraint that task \(k\) can
begin only once its dependencies have completed. The operator pipelines
independent dispatches and serializes dependent ones;
\(t_{\text{agent}}\) governs the constraint set.

The formal scheduling treatment and the full system-level extension
covering recursive task hierarchies are open problems (§7).

\section{7. Open Problems}\label{open-problems}

The framework opens several testable empirical questions:

\begin{enumerate}
\def\labelenumi{\arabic{enumi}.}
\item
  \textbf{Functional form of \(\rho(M)\)}. Is the dependence linear,
  logarithmic, sigmoidal? At what point do diminishing returns set in?
  Empirical measurement requires a controlled setting in which \(M\) can
  be varied while holding the task constant.
\item
  \textbf{Empirical measurement of \(c_p\), \(c_{i,j}\), \(c_r\)}. The
  framework defines per-channel cost scalars but does not assert their
  relative magnitudes or constancy across workflows. Field studies could
  measure how these scalars vary across task types, operator skill, and
  workflow design, and identify when paying earlier in a workflow is in
  fact cheaper than paying later. The cognitive psychology literature on
  task-switching cost [8, 9]
  is the natural anchor.
\item
  \textbf{Scheduling under dependencies for \(L_{\text{window}}\)}. The
  scheduling problem in §6 admits a dynamic programming formulation when
  the dependency graph is known. The harder problem is online scheduling
  with uncertainty about \(t_{\text{agent}}\).
\item
  \textbf{Measurement of \(I_{\text{novel}}\)}. The irreducible novelty
  of a task is currently a theoretical construct. One candidate
  operationalization is the conditional entropy of the task description
  given the agent's prior context. Empirical work requires committing to
  a definition.
\item
  \textbf{Multi-agent extensions}. When agents communicate with other
  agents, the analog of \(\rho\) involves no human bottleneck. The
  framework should extend, but the directional decomposition collapses
  and the cost scalars take on different values.
\item
  \textbf{Failure modes}. The framework assumes the task succeeds. When
  \(p_{\text{fail}} > 0\), review degenerates into another partial
  dispatch, and the formula's expectation involves a recursive term.
  Characterizing the breakdown regime is required for production
  deployment.
\item
  \textbf{Measurement of \(H_{\text{displaced}}\)}. The numerator is
  counterfactual: the human labor that would have been required without
  the agent. Two measurement strategies are available. The first is
  ensemble estimation, in which a panel of evaluators (human or model)
  with access to pre-training-scale priors estimates the labor cost of
  the task ex ante. The second is direct A/B timing of matched human and
  agent execution on the same task. Both have biases: ensemble
  estimation tends to overestimate when the agent's output is
  plausible-looking but subtly wrong, and A/B timing requires comparable
  task instances. Convergent estimates from both methods are the
  strongest signal.
\item
  \textbf{Recursive task hierarchies}. §6 introduces task-spawning as
  one of three amortization regimes for \(L_{\text{window}}\) but does
  not develop the recursive structure formally. A complete treatment of
  \(L_{\text{window}}\) under arbitrary task trees would account for
  meta-planning at parent tasks and the question of whether system-level
  leverage is bounded under recursion.
\item
  \textbf{Falsifying the directional asymmetry}. The framework's most
  distinctive operational claim is that \(\rho_{\text{in}}\) and
  \(\rho_{\text{out}}\) improvements produce phase-specific time savings
  governed by each phase's \(\alpha\) (§3.3). The values of \(\alpha\)
  for the studied phases are not known a priori and must be established
  as a precondition to the test, either by direct measurement (counting
  input vs output bits in observed exchanges) or by establishing a
  defensible ordering for the workflow under study. The framework's
  prediction depends only on the \emph{ordering} of \(\alpha\) across
  phases, not on specific values. A within-subjects falsification
  protocol: hold task class, agent capability, and operator constant;
  establish (or assume) that planning-phase \(\alpha\) exceeds
  review-phase \(\alpha\) for the studied tasks. Run two intervention
  conditions on matched task instances. The \(\rho_{\text{in}}\)
  condition deploys a faster human-to-agent input modality (e.g.,
  voice-to-text dispatch). The \(\rho_{\text{out}}\) condition deploys a
  denser agent-to-human output modality (e.g., a structured trace
  dashboard replacing prose summaries). For each condition, measure the
  per-phase time deltas \(\Delta t_{\text{planning}}\),
  \(\Delta \sum t_{\text{interrupt}_j}\), \(\Delta t_{\text{review}}\)
  relative to baseline. The framework predicts an interaction effect
  aligned with the \(\alpha\)-ordering: the \(\rho_{\text{in}}\)
  condition reduces time more in higher-\(\alpha\) phases, and the
  \(\rho_{\text{out}}\) condition reduces time more in lower-\(\alpha\)
  phases. The null result --- both interventions producing roughly
  proportional savings across all phases regardless of
  \(\alpha\)-ordering --- refutes the directional decomposition.
\item
  \textbf{Domain-localized memory transfer}. A second falsifiable claim
  sits in \(\rho(M)\): a properly-designed memory system raises \(\rho\)
  for tasks within the domain in which \(M\) was accumulated, without
  raising \(\rho\) for unrelated domains, and without negative transfer
  to adjacent ones. Protocol: accumulate \(M\) in domain \(A\) over a
  series of related tasks, then measure \(\rho\) on a held-out task in
  domain \(A\) and on a matched task in unrelated domain \(B\). The
  framework predicts \(\rho\) rises in \(A\) and is unchanged in \(B\).
  Two failure modes refute the framework's memory claim: every memory
  system either transfers globally (no domain awareness, \(\rho\) rises
  equally in \(A\) and \(B\), suggesting \(M\) is generic context rather
  than task-specific), or domain-localization comes at the cost of
  negative transfer (\(\rho\) rises in \(A\) but falls in \(B\)). Note
  that the test is contingent on whether such a system can be built; if
  no system exhibits the predicted pattern across many designs, the
  framework's account of memory is wrong even if the conservation law
  and directional decomposition stand.
\end{enumerate}

\section{8. Related Work}\label{related-work}

The supervisory control literature, beginning with Sheridan [1],
establishes that human supervision of autonomous systems carries a
measurable cost. The channel decomposition in §2 is one
operationalization of that cost; Sheridan's framework is qualitative and
orthogonal in axis (it concerns levels of decision authority rather than
directional information flow).

Common ground theory [10, 2] is the
qualitative ancestor of \(\rho(M)\). Clark and Brennan establish that
grounding cost decreases with shared context but do not quantify the
relationship in bits per second.

Mixed-initiative interaction [3] models the decision to
interrupt the human as a cost-benefit calculation. This anticipates the
per-channel cost scalars in §4.2 but treats interruption as binary
rather than as a continuous information channel.

Recent productivity research on generative AI [4, 5, 6] measures
aggregate output gains but does not provide a per-task unit of analysis.
The METR time-horizon work [7] measures the numerator
\(H_{\text{displaced}}\) but ignores the denominator.

Each component is anticipated in prior work; the unified ratio is not.

\end{document}